\documentclass{article}

\usepackage[final]{nips_2017}

\usepackage[utf8]{inputenc} \usepackage[T1]{fontenc}    \usepackage{hyperref}       \usepackage{url}            \usepackage{booktabs}       \usepackage{amsfonts}       \usepackage{nicefrac}       \usepackage{microtype}      

\usepackage{enumitem}
\usepackage{natbib,float}
\usepackage{color,algorithm,algorithmic}
\usepackage[disable]{todonotes}
\usepackage{amsfonts,amsmath,amssymb,amsthm,mathtools}
\usepackage{graphicx}
\usepackage{subfig}
\DeclareGraphicsExtensions{.pdf,.png,.gif,.jpg}

\title{Open-World Visual Recognition\\ Using Knowledge Graphs}

\author{
  Vincent P. A. Lonij, Ambrish Rawat, Maria-Irina Nicolae \\
  IBM Research -- Ireland\\
  Mulhuddart, Dublin 15, Ireland\\
  \texttt{\{vincentl, ambrish.rawat\}@ie.ibm.com, maria-irina.nicolae@ibm.com} \\
}

\begin{document}

\maketitle

\begin{abstract}

In a real-world setting, visual recognition systems can be brought to make predictions for images belonging to previously unknown class labels.
In order to make semantically meaningful predictions for such inputs, we propose a two-step approach that utilizes information from knowledge graphs.
First, a knowledge-graph representation is learned to embed a large set of entities into a semantic space.
Second, an image representation is learned to embed images into the same space.
Under this setup, we are able to predict structured properties in the form of relationship triples for any open-world image.
This is true even when a set of labels has been omitted from the training protocols of both the knowledge graph and image embeddings.
Furthermore, we append this learning framework with appropriate smoothness constraints and show how prior knowledge can be incorporated into the model.
Both these improvements combined increase performance for visual recognition by a factor of six compared to our baseline.
Finally, we propose a new, extended dataset which we use for experiments.
\end{abstract}

\section{Introduction} \label{sec:intro}
\makeatletter{}Natural images are simultaneously one of the richest sources of data about our environment and a challenging data type to analyze.
Building general purpose systems remains an unsolved paradigm of visual recognition.
However, it is critical for open-world applications to make meaningful predictions when confronted with inputs that are significantly different from the ones used during training.
In particular, these could be images from entirely new classes of objects.
To enable systems for making predictions in such scenarios, it is essential to incorporate additional knowledge.
Textual data is one such source that has widely been used for this purpose~\citep{devise_2013,socher2013zero}.
Models trained in these multi-modal settings mutually benefit from the shared information.
For instance, information from images can be used to discover new relationships between words.
Equivalently, semantic information from words can be used for image-related tasks like captioning or zero-shot learning.

In this work, we study an alternative source of information for visual recognition, namely knowledge graphs.
Before discussing the combined approach, we would like to distinguish between the following three scenarios of visual recognition:

\begin{enumerate}
	\item Standard classification setting: a model is faced with images that belong to classes that were available during training.

	\item Zero-shot setting: a model is expected to make predictions about images that do not belong to any of the classes seen at training time.
	However, side-information about the novel classes is traditionally available during training and used for making predictions.

	\item Open-world setting: encountered images belong neither to the seen classes, nor is any explicit side-information about their class is available at training time.
	We would like to point out that this case has not previously been studied.
\end{enumerate}

Knowledge graphs are a rich source of structured information. They are thus a natural choice for extracting semantic meaning about concepts.
A recognition system can subsequently be built on top of this information.
Owing to the underlying graph structure, these predictions can assume a more interpretable form as properties rather than class labels.
This capability to predict properties allows us to evaluate visual recognition systems across all of the aforementioned settings.
However, we believe that it is an essential aspect of visual recognition and provides novel capabilities.
It is worth noting that this combined framework also supports a \textit{knowledge-generation} scenario, where information from images can be utilized to extend the set of edges and nodes in a knowledge graph.
We discuss how this framework could be the foundation of a system that mimics human learning  by continually updating knowledge of the world based on sensory inputs.

We believe this work represents the first example of a method to use knowledge graph embedding for visual recognition in an open-world setting. The main innovations that enable this are the use of smoothing constraints in the graph embedding loss function, as well as an attention-based scheme to improve predictions of novel knowledge graph links.
Both these improvements combined increase performance for visual recognition by a factor of six compared to our baseline.
We also design a new dataset for evaluating the proposed tasks.

The rest of this paper is structured as follows. Section~\ref{sec:related} compares our approach to related methods.
Our proposed model is detailed in Section~\ref{sec:main}, where we explain the graph embedding (Section~\ref{sec:knowledge_graph_embedding}) and image embedding (Section~\ref{sec:image_embedding}) methods, as well as how to incorporate context in the prediction (Section~\ref{sec:context_model}).
The experimental study follows in Section~\ref{sec:experiments}.
Finally, we discuss the scope of this work and the relevant open questions in Section~\ref{sec:conclusion}.

\section{Related Work} \label{sec:related}
\makeatletter{}

The method proposed in this paper is situated at the crossroads of multiple domains, including semantic embedding, multi-relational knowledge bases, link prediction and zero-shot learning.
In this section, we position our work with respect to the most relevant approaches from these fields.

\paragraph{Semantic image embeddings}
A significant amount of work has been done to embed images into semantic vector spaces.
One line of research has focused on using text data to derive a semantic space for image embedding \citep{vendrov2015order}.
Deep visual-semantic embedding (DeViSE)~\citep{devise_2013} leverages images as well as unannotated text to provide a common representation.
The method uses a skip-gram model and a pre-trained visual model, both of which are embedded to a common space by learning a linear projection.
Our setup closely relates to~\citet{devise_2013}, as well as~\citet{socher2013zero} where they make class label predictions about unknown classes.
A joint sentence-image embedding was proposed in~\citet{kiros2014unifying} for machine translation.
In~\citet{xie2016image}, knowledge representations are learned with both triples and images from a joint objective.
However, they only consider knowledge generation for known entities, not open-world recognition.
This has been studied in the framework of zero-shot learning~\citep{palatucci2009zero}.
In our setup, we focus on the prediction of semantic relationships, as opposed to class labels for new images.
To our knowledge, visual recognition using such a structured semantic space has not explicitly been explored.
A parallel thread for predicting zero-shot classes is to use their attribute signatures~\citep{lampert09attributes,shi2014attributes,huang_learning_2015}.
Although this has proven to be a successful approach, it requires explicit attribute labeling of the images.
In our work, we are able to assign class labels and predict their properties without explicit annotation.

\paragraph{Knowledge graph embeddings}
Finding practical representations for knowledge graphs has been the focus of an important body of work.
In the standard setting, the algorithm is presented with triples encoding existing edges in the graph. Each triple $(e_h, r, e_t)$ links the head entity $e_h$ to the tail entity $e_t$ through the relation $r$ which holds true between them. A major direction of research for knowledge representation are translation models which embed information into a low-dimensional vector space.
TransE~\citep{bordes2013transe} is one of the simplest, yet most effective, formulations of this type.
The problem of knowledge graph embedding has also been tackled via tensor factorization for link prediction~\citep{nickel_rescal_2011, nickel_rescal_2012}, as well as latent factors models~\citep{sutskever_nips_2009, jenatton_nips_2012} and semantic energy matching~\citep{bordes_joint_2012, bordes_semantic_2014}.
In this work, we build on the neural tensor layer (NTL)~\citep{socher_nips_2013} which considers multiplicative mixing of entity vectors in addition to linear mixing.
We improve on this model by proposing a new objective function that enforces local smoothness on the scoring function.

\paragraph{Open-world visual recognition}
A general and more difficult framework has been formalized in~\citet{bendale2015towards} and~\citet{de2016online} as open-world recognition.
\citet{bendale2015towards} use an incremental approach to progressively include newly discovered classes in their model.
In this paper, we tackle the problem from the complementary direction of assigning semantic meaning to open-world images. We do this by making meaningful predictions for \textit{relationships} between open-world images and known entities.
Our experiments also demonstrate that effective prediction of these relationships can be achieved in a supervised setting.

\section{Knowledge Graph Embedding for Visual Recognition} \label{sec:main}
\makeatletter{}

\begin{figure}
\centering
  \includegraphics[width=0.9\textwidth]{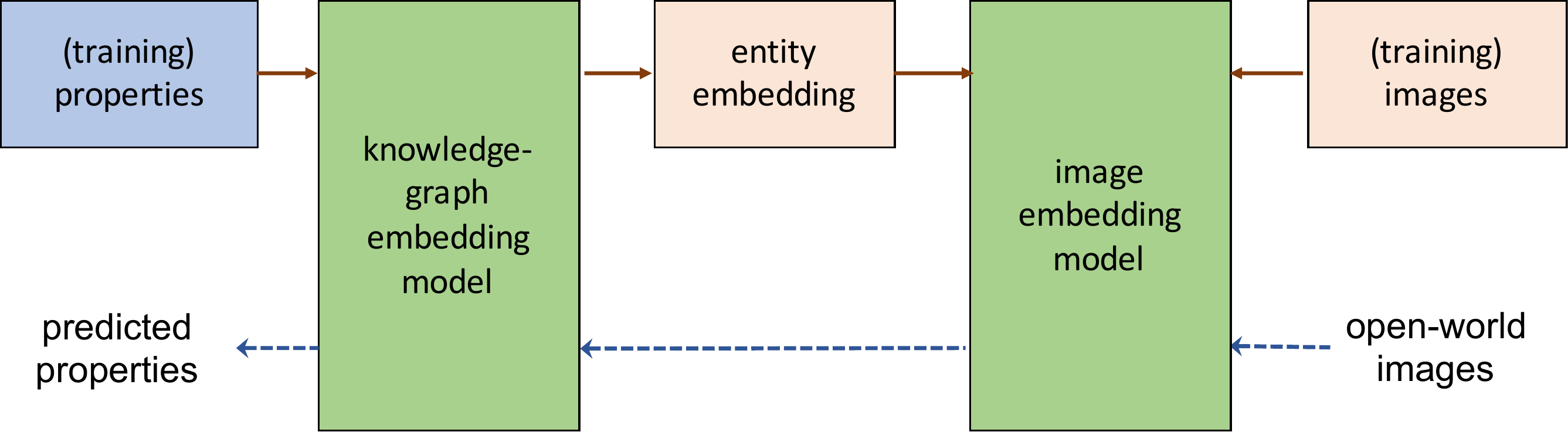}
\caption{Overview of our method. During training, data flows following the solid arrows. During open-world prediction, data flows from right to left following dashed arrows.}
\label{fig:overview}
\end{figure}

Our purpose is to find a semantic vector space in which entities from a knowledge graph and images can have a unified representation.
We arrive at this common space using a two-step approach.
First, a representation for the entities is learned using a knowledge-graph.
Subsequently, a map from images to vectors in that same space is learned.
This combination enables link-prediction for images. A high-level overview of our method is shown in Figure~\ref{fig:overview}.

To formalize these concepts we use the following notation.
We will refer to the set of real numbers as $\mathbb{R}$.
Scalars are denoted in lower case ($n$), vectors in bold ($\mathbf{v}$), while arbitrary sets in upper case symbols ($M$).
We define the hinge loss as $[t]_+ = \max(0, t)$, while for the $L_2$ norm we use the notation $||\cdot||_2$.
We use the set subtraction $A\setminus B$ to indicate the elements of $A$ that are not in $B$.

Consider the finite set of all knowledge, represented as a knowledge graph $\mathcal{G}$.
$\mathcal{G}$ can be modeled as a set of triples of the form $\{(e_i, r, e_j) \in T$, $e_i, e_j \in E, r \in R$\}, where $E$ is the set of entities in the vocabulary, and $R$ is the set of types of relations between entities.
All triples in $\mathcal{G}$ are taken to be true, while all triples not in $\mathcal{G}$ are false.
Having observed a subset $T^\prime \subset T$ of triples containing a subset of entities $E^\prime \subset E$, the task of link prediction is finding the triples in $T \setminus T^\prime$.
A standard version of this task addresses the setting where both $T'$ and $T \setminus T'$ are based on same entities $E^\prime$ and relationships $R$.
However, we will show that models can also be brought to predict triples for unknown entities $E\setminus E^\prime$.
To do this, we use information from a set of labeled images, $D := \{(x,e), x\in I, e\in E^\prime\}$
 where $I$ is the set of all images.
As mentioned above, this is done by learning an entity embedding $g: E \rightarrow V$ and an image embedding $h: I \rightarrow V$, where $V$ is a vector space in $\mathbb{R}^d$.
 We now describe how to determine $g$ and $h$.

\subsection{Knowledge Graph Embedding}
\label{sec:knowledge_graph_embedding}
\makeatletter{}
To train the knowledge embedding model, we have at our disposal a set of triples $T'$.
Let $g$ be the function mapping the entities to a vector space $g: E \rightarrow V$.
Now consider a scoring function $f$ defined over triples which attributes true triples low scores, and false triples high ones; for now, we assume it of general form $f : V \times R \times V \rightarrow \mathbb{R}$.
For a pair of entities $e_h$ and $e_t$ related with a relationship $r$, the score of the triple $(e_h,r,e_t)$ can be computed in different ways.
We adopt the bilinear form used in the neural tensor layer (NTL) architecture~\citep{socher_nips_2013}:
\begin{equation}
  \label{eq:triple_score}
  f(g(e_h), r, g(e_t)) = \tanh \left( \left( g(e_h)^T W^{[1:k]}_r \right) ^T g(e_t) + V_r \left[g(e_h) g(e_t) \right]^T + b_r\right),
\end{equation}
where $W^{[1:k]}_r \in \mathbb{R}^{d\times d \times k}$, $V_r \in \mathbb{R}^{k\times 2d}$ and $b_r \in \mathbb{R}^k$ are embedding parameters, $k$ is the number of slices as defined by~\citet{socher_nips_2013}, and $\left[g(e_h) g(e_t) \right]$ is the concatenation of the two vectors.
Attributing scores to triples allows us to use hinge rank loss for learning the embedding.
This margin-based loss function can then be written for the sample $T'$ as:
\begin{align}
\label{eq:original_loss}
  \mathcal{L}_E(g, f) &=  \frac{1}{|T'|} \sum_{(e_h, r, e_t) \in T', (e_h, r, e_t') \notin T'} \left[ \gamma + f(g(e_h), r, g(e_t)) - f(g(e_h),r,g(e'_t)) \right]_+,
\end{align}
where $\gamma$ is a margin parameter.
$(e_h, r, e_t')$ represents a false triple obtained by randomly replacing tail entity in $(e_h, r, e_t)$ with a different one.

A desirable property of a semantic space is that the scoring function $f$ should vary smoothly.
This is particularly relevant in the neighborhood of entity points.
We therefore modify Equation~\ref{eq:original_loss} by adding a noise term to the objective function:
\begin{equation}
  \label{eq:obj_kg}
  \mathcal{L}_S(g, f) = \alpha \mathcal{L}_E(g, f) + (1-\alpha) \mathcal{L}_E(\hat{g}, f),
\end{equation}
where $\alpha > 0$ is a trade-off parameter.
We use $\hat{g}$ to denote the operation of adding a normally distributed perturbation to the entity embeddings: $\hat{g}(e) = g(e)+\mathcal{N}(\boldsymbol{0},s\mathbf{I})$.
We will see that this is of particular importance when we use the embedding space for multi-modal transfer since the embedding vectors from, say, images will be near, but not exactly in the same place as corresponding entity vectors.

\subsection{Image Embedding} \label{sec:image_embedding}
\makeatletter{}
We model the image embedding $h: I \rightarrow V$ using a convolutional neural network~\citep{fukushima1980}.
For this task, we have access to a set $D$ of labeled images $(x, e)$, where the labels $e \in E$ are entity types from the knowledge graph.
Let $g^*$ be the learned entity embedding $g$ obtained by minimizing the objective in Equation~\ref{eq:obj_kg}.
In order to embed images into the previously obtained representation space, we propose to use a least-squares objective mapping images onto their corresponding entities:
\begin{equation}
  \label{eq:obj_image}
  \mathcal{L}_I(h) = \frac{1}{|D|} \sum_{(x,e) \in D} || h(x)- g^*(e)||^2_2.
\end{equation}
To predict knowledge graph links for images in open the world settings we first compute $h(x), x \in I$.
Next, we will describe how the resulting image embedding vector $\mathbf{v}$ can then be used to predict true triples.

\subsection{Context Heuristic} \label{sec:context_model}
\makeatletter{}For the task of link prediction, we consider the scores of a set of possible links, $L \subset R\times E$.
For an image $x$, this can be computed by $\{f(h(x),r,g(e)), (r,e) \in L\}$.
It is important to note that the set of true knowledge $\mathcal{G}$ is finite and significantly smaller than the set of all possible triples.
Conventional approaches to embeddings do not account for this skewness.
We address the problem by re-scoring the triples using a context heuristic.
For an image $x_0$ and its link $(r_0, e_0)$,

$$ u(x_0, r_0, e_0) = \frac{a \cdot b_1}{a \cdot b_1 + (1-a) \cdot b_2},$$
where
\begin{align}
a &= \frac{1}{|E'|} \sum_{e\in E'}\mathbb{I}[(e,r_0,e_0)\in T'], \\
b_1 &= \frac{1}{\sqrt{2 \pi \sigma_{\textrm{true}}^2}} \exp \left(\frac{-[ f(h(x_0),r_0,t_0) - \mu_{\textrm{true}}]^2}{2\sigma_{\textrm{true}}^2} \right),\\
b_2 &= \frac{1}{\sqrt{2 \pi \sigma_{\textrm{false}}^2}} \exp \left(\frac{-[f(h(x_0),r_0,t_0) - \mu_{\textrm{false}}]^2}{2\sigma_{\textrm{false}}^2} \right).
\end{align}

This heuristic is motivated from a probabilistic intuition where prior information about the context is propagated for re-scoring a triple.
$a$ can then be seen as the attention for a link $(\cdot, r_0, e_0)$.
Similarly, $b_1$ and $b_2$ capture the information about true and false triples that was learned in $f$.
The estimates for $\mu_{\textrm{true}}$ and $\sigma_{\textrm{true}}$ (and similarly, $\mu_{\textrm{false}}$ and $\sigma_{\textrm{false}}$) are computed as the mean and standard deviation of the scores for the set $T'$ and respectively a false sample $\hat{T} \subset E \times R\times E \setminus T$.

\section{Experiments} \label{sec:experiments}
\makeatletter{}In this section, we provide an experimental study substantiating the useful properties of our method.
We start by introducing WN1M, a new knowledge dataset which we use throughout our experiments.
We evaluate the performance of our method for multiple tasks, comparing different settings to show their usefulness.
Additionally, we explore the learned embedding space in terms of visualizations of property regions and entity discrimination capacities.

\subsection{Data Sources}
\label{sec:data}

To demonstrate the capability of our method to do visual recognition in an open-world setting, we craft a knowledge dataset with a larger number of triples.
We create a new set of true triples based on the WordNet knowledge graph by reasoning over the links and expanding transitive properties up to depth four.
We do this to ensure that the semantic embedding accounts for transitivity.
We select only triples where both head and tail are nouns and select the relationships (\textit{hypernym, hyponym, part meronym, part holonym, member meronym, member holonym}).
This process roughly yields $10^6$ triples, hence we call this dataset WordNet 1 million (WN1M).
The triples are split into three disjoint sets to be used for the tasks discussed in Section~\ref{sec:intro}.

\begin{enumerate}
  \item A training set containing 1 million triples based on entities $ e \in E^\prime$,
  \item A standard test set containing 20,000 triples based on entities that also occur in the training set ($e \in E^\prime$),
  \item A hard test set containing all the triples associated with a selected set of 50 entities $e \in E \setminus E^\prime$  that we held out from sets 1 and 2 and for which images are available.
\end{enumerate}

To train the image embedding, we use the ILSVRC-2012 image dataset~\citep{ILSVRC15}, which we split into four distinct sets:

\begin{enumerate}
  \item A training set containing images from 750 classes,
  \item A validation set (VAL) with distinct images from the same 750 classes,
  \item A zero-shot test set (ZS) of 200 image classes that are held out entirely during the training of the image model, but are present in the knowledge graph,
  \item An open world test set (OW) of 50 classes that are left out from the image embedding training corresponding and correspond to the 50 entities left out from the KG embedding training.
\end{enumerate}
We would like to point out that the classes for each of the above categories were picked randomly and fixed for all the experiments. 
Following the protocol of \citet{xian_zeroshot_2017}, we also evaluate our method on the 1K most populated classes of ImageNet that are not part of ILSVRC-2012. These 1K entities are present in the KG, therefore this set falls in the ZS category.

\subsection{Experimental Setup}

\paragraph{Knowledge embedding}
We train two knowledge graph embedding functions, one based on the original NTL architecture~\citep{socher_nips_2013}, and one based on our adapted smooth NTL method (SNTL, Equation~\ref{eq:obj_kg}).
For both embedding algorithms we choose the dimensionality of the embedding space  $d=60$, with $k=6$ slices in the layer.
Both embeddings are trained with batch optimization for 300 epochs.
The batches contain 10,000 triples and are generated randomly from the training set.
A corrupted triple is generated for each true triple by replacing the tail entity with a random one.
The Gaussian noise parameter is set to $s=0.1$, slightly larger than the error obtained from the image embedding models (Equation \ref{eq:obj_image}) in each dimension.

\paragraph{Image embedding}
For the image model, we use the VGG16 architecture ~\citep{Simonyan2014vgg}. Following \citet{devise_2013}, we use convolution-filter weights which were pre-trained for a classification task.
In the dense layers we use exponential linear units and dropout layers that randomly set 30\% of values to zero.
The final layer has a number of units equal to the dimensionality of the knowledge graph embedding.
The output is normalized to have unit $L_2$ norm.
We use the RMSProp optimization procedure~\citep{tieleman2012lecture} to train two separate image embedding models, one with the NTL entity vectors as targets which we call Image NLT (INTL) and one with the smooth SNTL entity vectors as targets (SINTL).

\subsection{Performance Evaluation}
\label{sec:performance}

\begin{table}
\centering
\caption{Comparison of knowledge graph embedding models without images. Mean rank ($\mu_r$) is reported as the fraction of the total number of incorrect triples.}\label{tab:standard_link_prediction}
\begin{small}
  \begin{tabular}{lrrr}
\toprule
            & TransE & NTL & SNTL \\
\midrule
     $\mu_r$ & 0.0048 &0.0030 & 0.0028 \\
\bottomrule

\end{tabular}
\end{small}
\end{table}

Two metrics that are commonly used to quantify the precision of a link prediction algorithm are: (a) the mean rank of true triples in a list of possible triples sorted by the scoring function, and (b) the fraction of true triples that rank in the top 10. These metrics are computed by first appending a true triple to the list of all incorrect triples and then sorting this list according to the scoring function $f$.
We argue that these metrics are less meaningful for the task of visual recognition because they measure the algorithms ability to extract \textit{all} knowledge from one image. Instead, it is more relevant to report how well at least some previously unknown knowledge can be extracted from an image.
In the following, we still report mean ranks in order to compare to related work.
However, we argue that two more meaningful metrics are: (a)  the number of true triples in the top $n$ of a sorted list of all triples with the same head ($t@n$), and (b) the fraction of images for which at least one true triple is ranked in the top $n$ ($f@n$).

\paragraph{Link prediction}
As a consistency check, we compute the performance of our method on the link prediction task using three different knowledge graph embedding methods: TransE~\citep{bordes2013transe}, NTL and SNTL (Table~\ref{tab:standard_link_prediction}).
Both NTL-based methods outperform TransE which can be explained by the fact that the geometric constraints in TransE make it incapable of modeling transitive relationships.
We observe that the additional constraint of smoothness does not significantly impact the performance for this task.
We will now show how SNTL significantly outperforms NTL on visual recognition tasks.

\paragraph{Image recognition}

\begin{figure}

\centering
\subfloat{
  \includegraphics[width=0.49\linewidth]{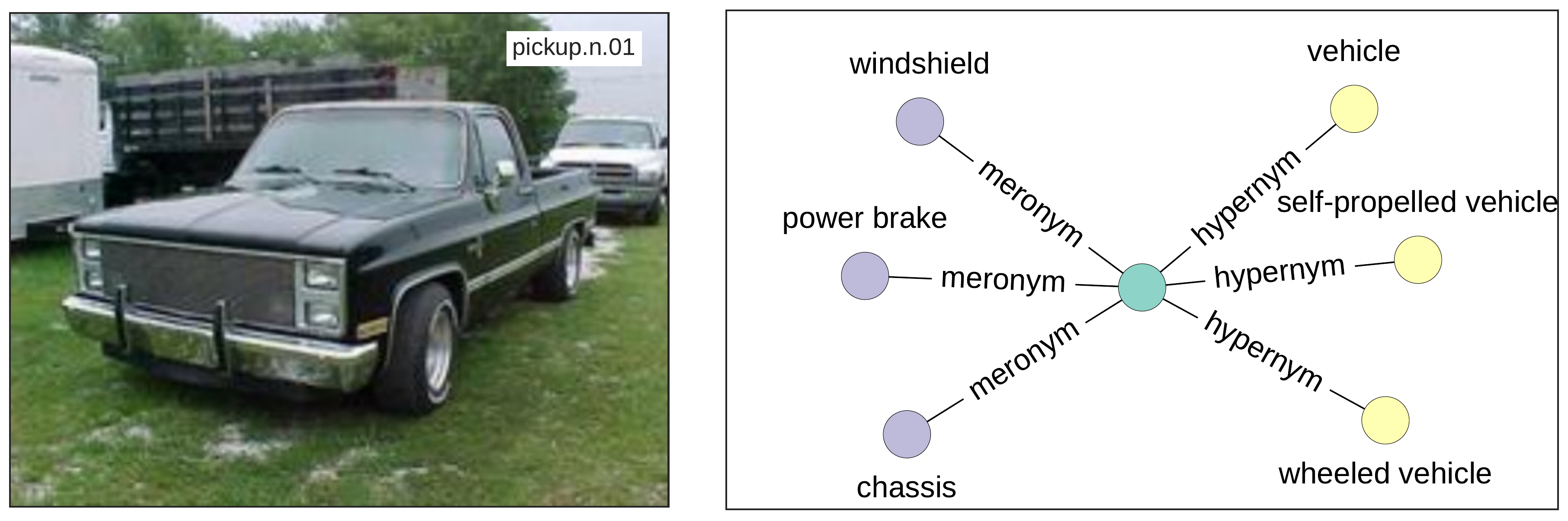}
}
\subfloat{
  \includegraphics[width=0.49\linewidth]{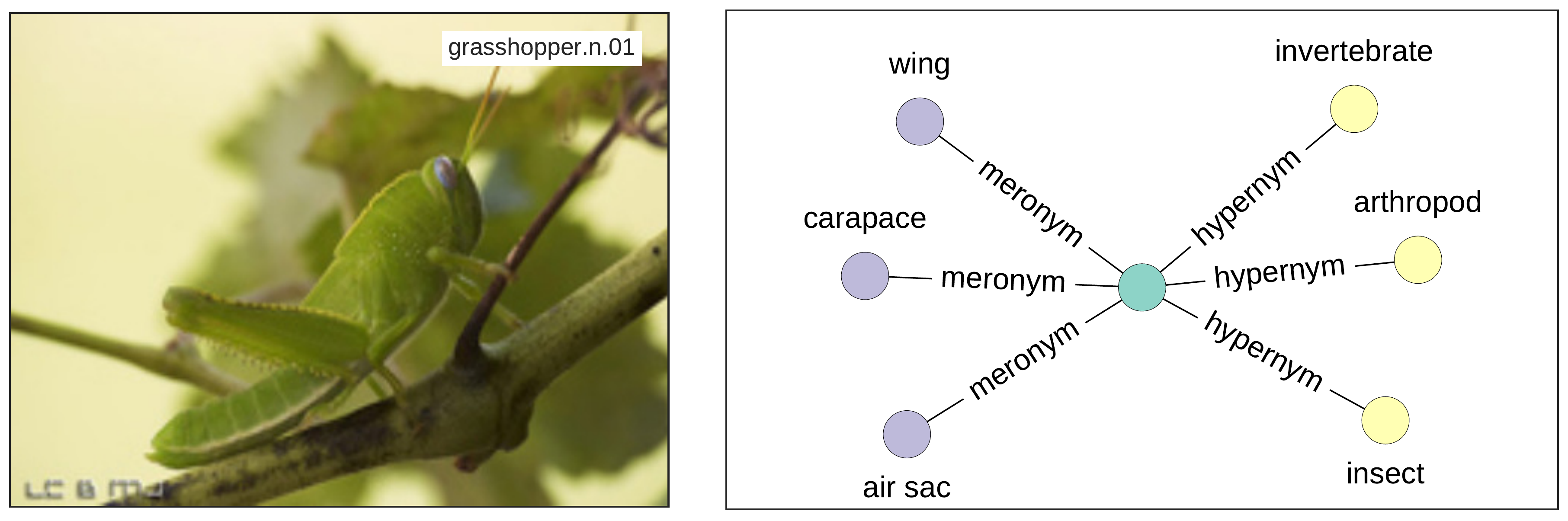}
}
\hspace{0mm}
\subfloat{
  \includegraphics[width=0.49\linewidth]{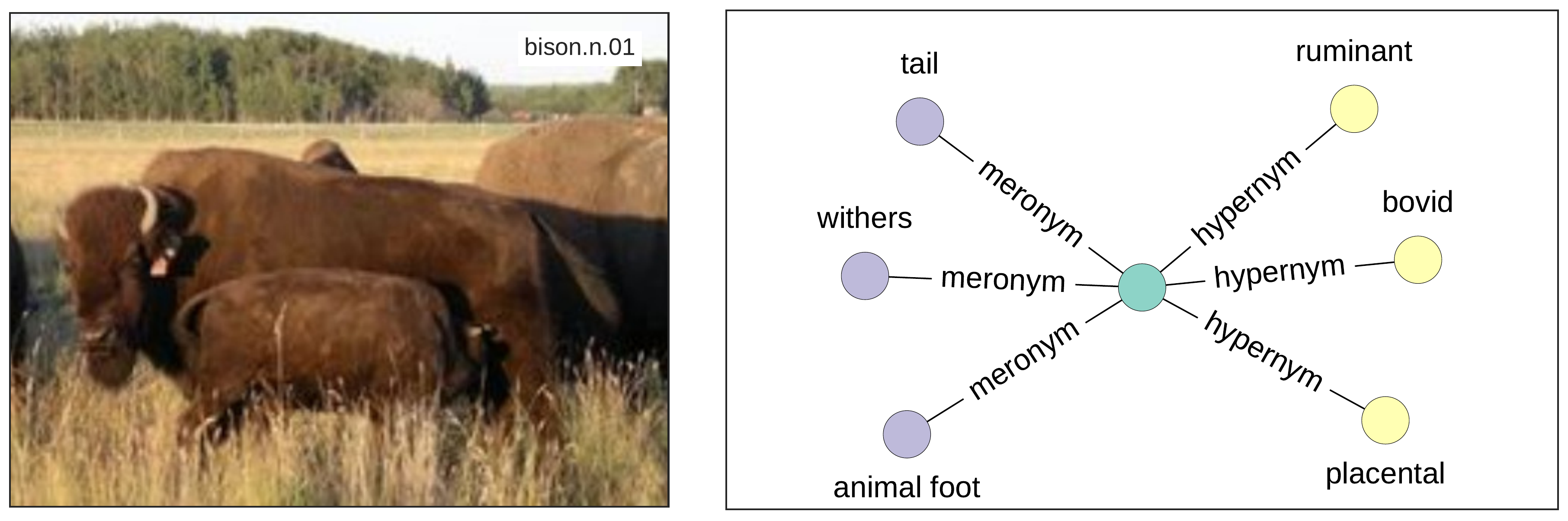}
}
\subfloat{
  \includegraphics[width=0.49\linewidth]{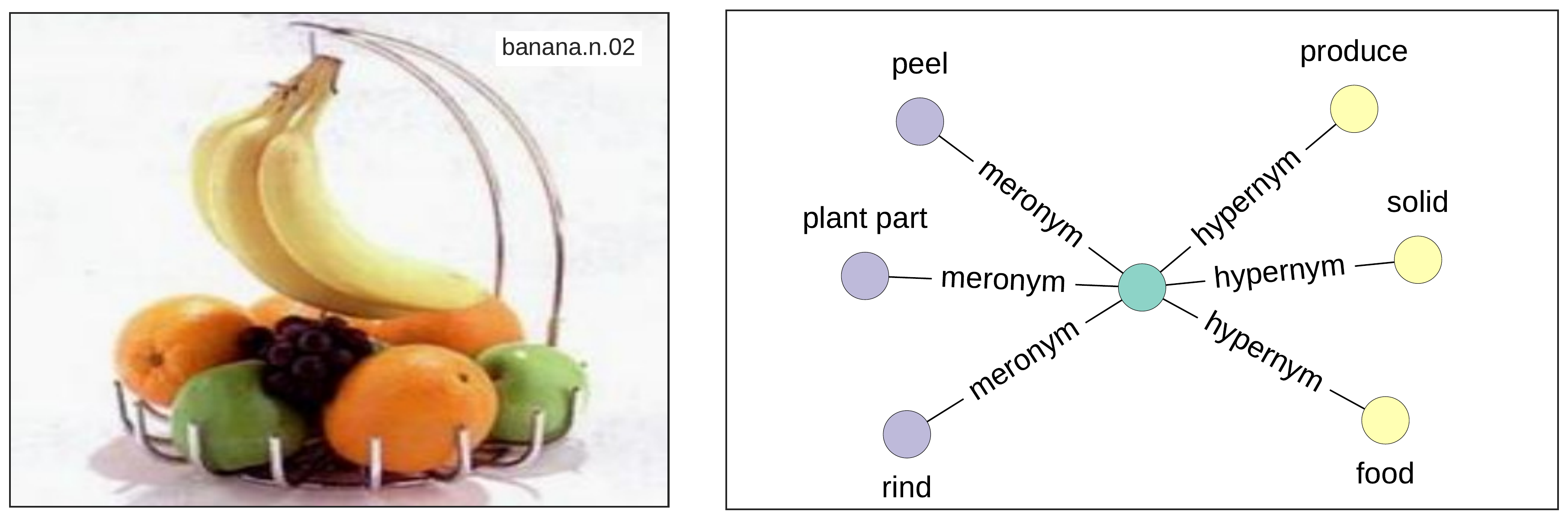}
}
\caption{Examples of knowledge graph outputs. Top three hypernyms (A is a kind of B) and top three meronyms (A has part B) are shown.
The top row of images is obtained in a zero-shot setting, while the bottom row in the open-world setup.}
\label{fig:image_examples}
\end{figure}

We compare the results of the base (INTL) and smooth (SINTL) models for image link prediction, both with and without the context heuristic and compare results on three different datasets defined in Section~\ref{sec:data}. Table~\ref{tab:per_image} shows that of the top 3 triples, on average about 2 are correct for our best model on the zero-shot (ZS) data. The open-world (OW) case, where classes were also omitted from the knowledge graph performs only slightly worse than the zero-shot (ZS) dataset. Performance on the more challenging 1K dataset is lower.
This is consistent with the findings of~\citet{xian_zeroshot_2017}, and  is likely due to the fact that several classes in the 1K set share no properties with classes in ILSVRC-2012 image dataset.
Across all datasets, our SINTL model significantly outperforms the baseline INTL model and the use of our context heuristic yields improvement for both INTL and SINTL models.
Figure~\ref{fig:image_examples} shows selected examples of images with their predicted links.

Next, we turn our attention to analyzing collections of images.
Table~\ref{tab:class_mean} shows the result of first collecting all image embedding vectors for the same image class and then performing link prediction using the vector mean. The results in this table are consistently better than on a per-image basis, which indicates that a set of images known to belong to the same class can be used to generate an accurate knowledge graph representation of the object class.
We can thus use our model to extend the original graph by adding entities whose links are determined based on image data alone.

\begin{table}
  \caption{Property prediction performance per image. The table reports the fraction of images with at least one correct triple in the top 3 ($f@3$), the mean rank of true triples as a fraction of the total number of triples ($\mu_r$), and the average number of true triples in the top 3 ($t@3$). }
  \label{tab:per_image}
  \centering
\begin{small}
\begin{tabular}{lrrrrrrrrr}
\toprule
Dataset                 &   \multicolumn{3}{c}{ZS}  & \multicolumn{3}{c}{OW} & \multicolumn{3}{c}{1K} \\
Metric                  &   $f@3$  &   $\mu_r$ & $t@3$ &   $f@3$  & $\mu_r$ & $t@3$ &   $f@3$  & $\mu_r$ & $t@3$ \\
\midrule
Random           &    0.003 & 0.500    &0.003    &0.001    &0.500       &0.001&   0.003&         0.500&     0.003\\
 INTL            &    0.161 &   0.082 &    0.276 &    0.144 &   0.148 &    0.285 &    0.101 &   0.207 &    0.182 \\
 INTL + Context  &    0.692 &   0.067 &    1.358 &    0.562 &   0.127 &    1.031 &    0.456 &   0.174 &    0.769 \\
 SINTL           &    0.316 &   0.061 &    0.523 &    0.269 &   0.114 &    0.471 &    0.144 &   0.183 &    0.234 \\
 SINTL + Context &    \textbf{0.855} &   \textbf{0.044} &    \textbf{1.986} &    \textbf{0.756} &   \textbf{0.091} &    \textbf{1.702} &    \textbf{0.632} &   \textbf{0.145} &    \textbf{1.326} \\
    \bottomrule
\end{tabular}
\end{small}
\end{table}

\begin{table}
  \caption{Property prediction performance per class. The table reports the same metrics as Table~\ref{tab:per_image}.}
    \label{tab:class_mean}
  \centering
\begin{small}
\begin{tabular}{lrrrrrrrrr}
\toprule
Dataset                 &   \multicolumn{3}{c}{ZS}  & \multicolumn{3}{c}{OW} & \multicolumn{3}{c}{1K} \\
Metric                 &   $f@3$  &   $\mu_r$ & $t@3$ &   $f@3$  & $\mu_r$ & $t@3$ &   $f@3$  & $\mu_r$ & $t@3$ \\
\midrule
 INTL            &    0.154 &   0.061 &    0.241 &    0.167 &   0.125 &    0.250 &    0.066 &   0.191 &    0.095 \\
 INTL + Context  &    0.810 &   0.049 &    1.585 &    0.688 &   0.106 &    1.188 &    0.547 &   0.162 &    0.881 \\
 SINTL           &    0.400 &   0.043 &    0.687 &    0.375 &   0.080 &    0.667 &    0.176 &   0.171 &    0.249 \\
 SINTL + Context &    \textbf{0.928} &   \textbf{0.031} &    \textbf{2.174} &    \textbf{0.812} &   \textbf{0.059} &    \textbf{1.917} &    \textbf{0.671} &   \textbf{0.136} &    \textbf{1.472} \\
\bottomrule
\end{tabular}
\end{small}
\end{table}

\subsection{Properties of the Semantic Space}
We now study the properties of the semantic representation space learned by the proposed models through visualizations, as well as a relevant metric.

\begin{figure}
\centering

\subfloat{
  \includegraphics[width=0.55\textwidth]{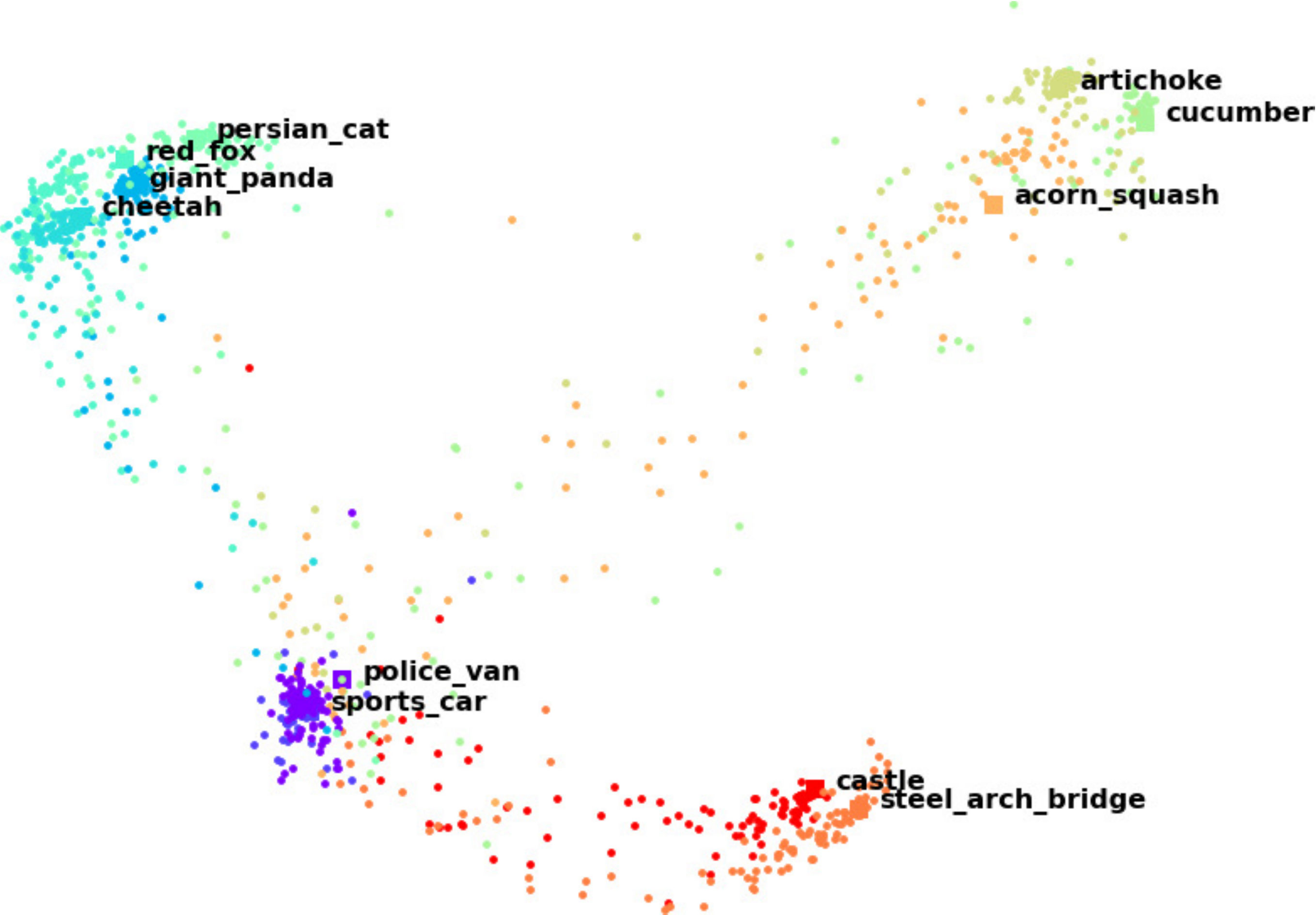}
  }
\subfloat{
 \includegraphics[width=0.45\textwidth]{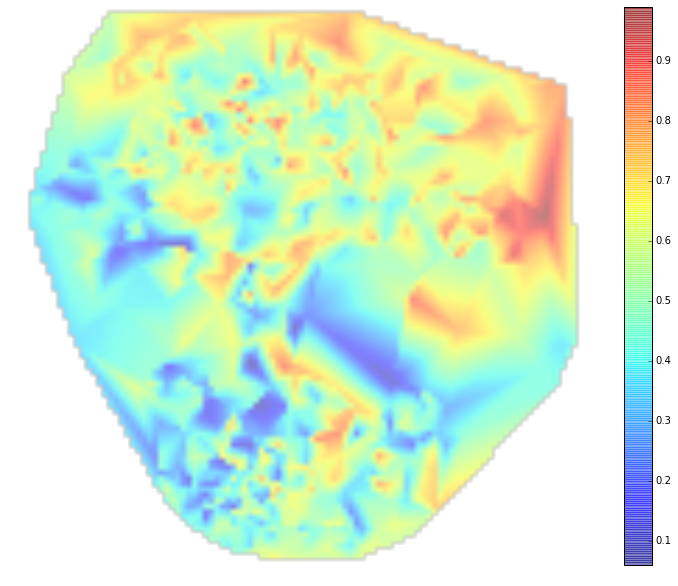}
}
\caption{Visualization of the semantic space. Left: image  and entity vectors obtained from $g$ and $h$.
Right: The scoring function varying smoothly for the property ``is a kind of mammal''.}
\label{fig:score_surface}
\end{figure}

\paragraph{Entity discrimination} In this experiment, we visualize the spatial distribution of entities belonging to different categories.
Figure~\ref{fig:score_surface} (left) maps a subset of entities and their corresponding images to the embedding space, which is represented in the figure after 2D PCA ($\sim$55\% of the original variance explained).
The points in the plot represent actual images, and their color corresponds to their true entity.
The classes present in the figure are part of different categories (vegetables, animals, structures and vehicles).
We observe that the entities are grouped together according to their parent category, showing that the embedding has semantic properties and the capacity to cluster similar entities.

\paragraph{Property regions}
In addition to a semantic meaning of entities, the feature space $V$ also incorporates semantics with respect to knowledge graph links.
To visualize the properties of the scoring function $f$ over $V$, we compute its values for a set of triples $(\mathbf{v}, 'hypernym', 'mammal')$, where $\mathbf{v}$ is a randomly sampled set of vectors in $V$.
Figure~\ref{fig:score_surface} (right) presents the results after applying PCA and retaining the most important two dimensions (which explain $\sim$25\% of the variance in the feature space).
Notice that the scoring function varies smoothly, especially for low values of $f$ (in blue), which represent the predicted true triples.

\paragraph{Smoothness}
As shown in Section~\ref{sec:performance}, smoothness of $f$ greatly improves visual recognition.
We can quantify smoothness of the semantic embedding by estimating its Lipschitz continuity constant $\ell$: the lower the value, the smoother the function.
This measure quantifies the largest variation of a function under a small change in its input.
In practice, we are unable to compute this theoretical metric for our model.
Instead, we propose to estimate $\ell$ as an average value over the training set using the difference in loss function between each triple and its noisy version.
We find that our modified loss function yields $\ell= 0.174$ in contrast with $\ell= 0.645$ for the NTL model, proving that our semantic embedding is smooth.

\section{Conclusion and Outlook} \label{sec:conclusion}

We presented a mutual semantic embedding space based on knowledge graphs for visual recognition.
Our system extends beyond label prediction, as we achieve knowledge extension through link prediction.
This makes the method ideally suited for open-world visual recognition where inputs often lie outside of what was anticipated by design.
The results we report show an accuracy level of practical use.
Furthermore, the output of our method  can be used in reasoning systems to facilitate automated decision-making based on images.

Our work also indicates the need for further research in a few different areas.
The fact that a context heuristic can improve accuracy of link prediction suggests that existing graph embedding methods do not learn the distribution of relationship-tail pairs in the training set.
This suggests the need to develop ways to incorporate this prior knowledge during the learning process in a principled way.
Our method does require that collections of images are provided that are known to belong to the same class.
Therefore, a natural extension would be to combine our method with algorithms that can determine if two images belong to the same class.

One particularly significant feature of our method is that it treats on equal footing both visual recognition (for known and unknown classes), as well as knowledge graph extension.
The method could thus be the foundation of a self learning system that updates its knowledge based on visual stimuli and automatically improves its recognition capabilities.

\begin{footnotesize}
\bibliography{references}
\end{footnotesize}

\end{document}